\documentclass[runningheads]{llncs}
\usepackage[pdftex, pagebackref=true, pdfpagelabels=true,
hypertexnames=true, plainpages=false, naturalnames=false,
bookmarks=true, bookmarksnumbered=true,
pdfpagemode=UseOutlines,]{hyperref}

\usepackage[numbers]{natbib}

\newcommand{\repeatthanks}{\textsuperscript{\thefootnote}}

\usepackage{graphicx}\graphicspath{ {figures/} }
\usepackage{multirow}
\usepackage{booktabs, microtype}
\usepackage{multicol}

\begin{document}
\title{Deep Learning for Suicide and Depression Identification with Unsupervised Label Correction}
\titlerunning{Suicide vs Depression Classification with Noisy Labels}
%
\author{Ayaan Haque\inst{1}{}\thanks{Equal Contribution} \and
Viraaj Reddi\inst{1}{}\repeatthanks \and
Tyler Giallanza\inst{2}{}
}

\authorrunning{Haque et al.}
%
\institute{Saratoga High School, Saratoga, CA, USA \and
Department of Psychology \& Princeton Neuroscience Institute, Princeton University, Princeton, NJ, USA 
}
\maketitle              
\begin{abstract}
Early detection of suicidal ideation in depressed individuals can allow for adequate medical attention and support, which can be life-saving. Recent NLP research focuses on classifying, from given text, if an individual is suicidal or clinically healthy. However, there have been no major attempts to differentiate between depression and suicidal ideation, which is a separate and important clinical challenge. Due to the scarce availability of EHR data, suicide notes, or other verified sources, web query data has emerged as a promising alternative. Online sources, such as Reddit, allow for anonymity, prompting honest disclosure of symptoms, making it a plausible source even in a clinical setting. However, online datasets also result in inherent noise in web-scraped labels, which necessitates a noise-removal process to improve performance. Thus, we propose SDCNL, a deep neural network approach for suicide versus depression classification. We utilize online content to train our algorithm, and to verify and correct noisy labels, we propose a novel unsupervised label correction method which, unlike previous work, does not require prior noise distribution information. Our extensive experimentation with various deep word embedding models and classifiers display strong performance of SDCNL as a new clinical application for a challenging problem. \footnote{We make our \textbf{supplemental}, dataset, web-scraping script, and code (with hyperparameters) available at \href{https://github.com/ayaanzhaque/SDCNL}{https://github.com/ayaanzhaque/SDCNL}}

\keywords{Suicide/Depression \and Noisy Labels \and Deep Learning \and Online Content \and Natural Language Processing \and Unsupervised Learning}
\end{abstract}
\section{Introduction}
\label{sec:introduction}

Depression remains among the most pressing issues worldwide and can often progress to suicidal ideation or attempted suicide if left unaddressed. Diagnosis of depression and identification of when it becomes a risk of attempted suicide is an important problem at both the individual and population level. Many existing methods for detecting suicidal ideation rely on data from sources such as questionnaires, Electronic Health Records (EHRs), and suicide notes \cite{ji2020suicidal}. However, acquiring data in such formats is challenging and ultimately results in limited datasets, complicating attempts to accurately automate diagnosis.

Conversely, as the Internet and specifically social media have grown, online forums have developed into popular resources for struggling individuals to seek guidance and assistance. These forums have potential to be scraped to create datasets for automated systems of mental health diagnosis, as they are extensive and free to access. Especially for neural network based approaches that require large datasets to be trained efficiently, a growing number of studies are using this data for diagnostic purposes, which are detailed in this review paper \cite{ji2020suicidal}.

In particular, Reddit has emerged as an important data source for diagnosing mental health disorders \cite{inproceedings}. Reddit is an online social media forum in which users form communities with defined purposes referred to as subreddits. Certain subreddits discuss dealing with mental health and openly explain their situations (r/depression and r/SuicideWatch). Reddit specifically allows users to create alternate and discardable accounts to ensure privacy and anonymity, which promotes disclosure and allows those with little support systems in real life to receive support online \cite{de2014mental}. The wide user base, honesty of these online settings, and moderated screening of these posts to ensure legitimacy provides an unprecedented opportunity for computationally analyzing mental health issues on a large scale.

Despite the extensive research into classifying between healthy and mentally unstable patients through text, there remains little work focused on detecting when individuals with underlying mental health struggles such as depression are at risk of attempting suicide. This represents an important clinical challenge, both for the advancement of how depression is treated and for implementing interventions \cite{bering2018suicidal,leonard1974depression}. Distinguishing between suicidality and depression is a more fine-grained task than distinguishing between suicidal and healthy behavior, explaining the lack of current solutions. Online data has traditionally been difficult to use in such fine-grained situations, because labels for such data are often unreliable given their informal nature and lack of verification. In particular, labeling data based on subreddit relies on self-reporting, since each user chooses which subreddit they feel best reflects their mental state; thus, they may over or under report their diagnosis. This concept is referred to as \emph{noisy labels} as there is a potential for certain labels to be corrupted. Estimates show that noisy labels can degrade anywhere from 10\% to 40\% of datasets \cite{song2020learning}, presenting serious challenges for machine learning algorithms.

Current attempts to address the noisy label problem can be categorized into three notable groups: noise-robust methods, noise-tolerant methods, and data cleaning methods \cite{song2020learning, frenay2014survey}. Noise-robust approaches rely on algorithms that are naturally less sensitive to noise (e.g. lower dimensional or regularized algorithms), whereas noise-tolerant methods directly model the noise during training. Although both approaches have received considerable attention in the image-processing domain \cite{song2020learning}, these methods do not transfer to NLP algorithms. In the NLP domain, there have been a few recently proposed noisy label methods \cite{NEURIPS2018_ad554d8c, jindal2019effective, zheng2019meta}. However, the proposed methods have limitations for our task. For example, some methods utilize a smaller set of trusted data to correct a larger set of noisy data \cite{NEURIPS2018_ad554d8c, zheng2019meta}, which is infeasible for our application as there is no way to evaluate which posts are accurate. Other methods require training a network directly and end-to-end from corrupted labels \cite{jindal2019effective}, which both requires a relatively large amount of data and is less capable of leveraging transfer learning from pre-trained, state-of-the-art models \cite{devlin-etal-2019-bert, reimers2019sentencebert, cer2018universal}.

Data cleaning methods are more suitable for the present task. However, most existing label cleaning methods make assumptions about or require knowledge on the distribution of noise in the dataset \cite{jiang2020learning, hendrycks2018using}. In our use-case, where there is no prior knowledge of the noise distribution, an unsupervised method, such as clustering, is required. Although there are a few methods that use unsupervised clustering algorithms for noisy label learning \cite{bouveyron:hal-00325263}, none of these correct labels. Rather, they train a model to be robust to noise through instance weighting or exclusion. These methods would be problematic for our task; due to the high noise proportion, weighting or removing a high volume of data would damage performance, especially for deep neural networks require large amounts of data. Thus, the present task requires an unsupervised method for data cleaning that utilizes label correction rather than elimination. To the best of our knowledge, there are no current methods which perform label correction using unsupervised clustering methods, and particularly not in the NLP domain.

In this paper, we present SDCNL to address the unexplored issue of classifying between depression and more severe suicidal tendencies using web-scraped data and neural networks. Our primary contributions can be summarized as follows:

\begin{itemize}
    \item Deep neural network sentiment analysis applied for depression versus suicidal ideation classification, an important but unexplored clinical and computational challenge
    
    \item A novel, unsupervised label correction process for text-based data and labels which does not require prior noise distribution information, allowing for the use of mass online content
    
    \item Extensive experimentation and ablation on multiple datasets, demonstrating the improved performance of all SDCNL components on the challenging proposed task
\end{itemize}


\section{Methods}
\label{sec:methods}

\begin{figure}
    \centering
    \resizebox{0.65\linewidth}{!}{
    \includegraphics[width = \linewidth]{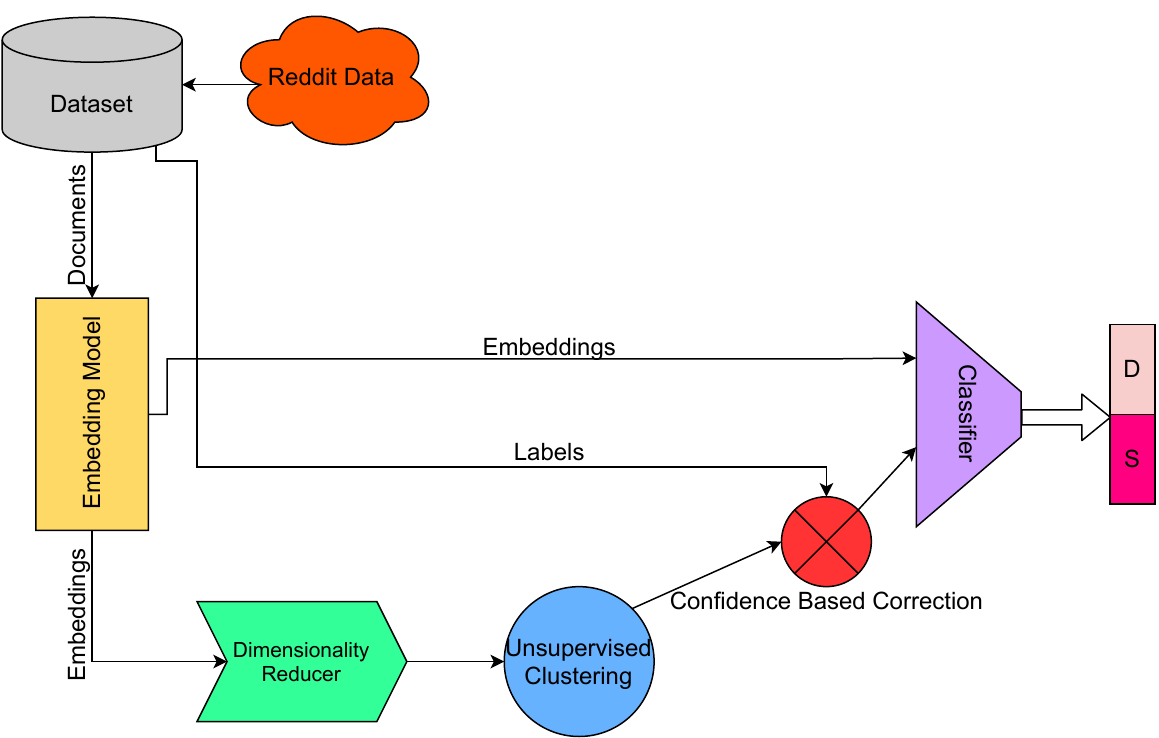}
    }
    \caption{Schematic of the SDCNL pipeline used for classification of suicide vs depression and noisy label correction via unsupervised learning.}
    \label{fig:pipeline}
\end{figure}

The SDCNL method is outlined in Figure \ref{fig:pipeline}. We begin by processing text data scraped from Reddit with word embedding models. These embeddings are then processed with an unsupervised dimensionality reduction algorithm, as clustering algorithms do not perform well in high-dimensional domains \cite{steinbach2004challenges}. The reduced embeddings are then inputted into a clustering-based algorithm which predicts new labels in an unsupervised manner, meaning it is independent of noise. These alternate labels are compared against the ground-truth labels using a confidence-based thresholding procedure in order to correct the ground-truth labels. The corrected set of labels are then used to train a deep neural network in a supervised fashion.


\subsection{Embedding Models}

Our framework initially utilizes word embedding models to convert raw documents, which in our case are referred to as posts, to numerical word embeddings. Our proposed method can be used with any embedding models, but given our task, we require greater-than-word text embedding models optimized to work with phrases, sentences, and paragraphs. We experiment with 3 state-of-the-art transformers: Bidirectional Encoder Representations from Transformers (BERT) \cite{devlin-etal-2019-bert}, Sentence-BERT \cite{reimers2019sentencebert}, and the Google Universal Sentence Encoder (GUSE) \cite{cer2018universal}. BERT is a state-of-the-art, bidirectionally trained transformer that achieves high performance on various benchmark NLP tasks, and outputs a 768 $\times$ 512 dimensional vector of embeddings. Sentence-BERT is an extension of the original BERT architecture that is retrained and optimized for longer inputs and better performance during clustering, and it outputs a 768 $\times$ 1 dimensional vector. GUSE is a transformer also trained and optimized for greater-than-word length text, but rather returns a 512 $\times$ 1 dimensional vector. 

Some classifiers require word level representations for embeddings, while others require document level representations. BERT outputs both multi-dimensional word level embeddings as well as document level embeddings, which are provided by CLS tokens. Depending on what the classifier requires, we vary the inputted embeddings to match the classifier's requirement. In addition, we also experiment on three vectorizers as baselines: Term Frequency–Inverse Document Frequency (TFIDF), Count Vectorizer (CVec), and Hashing Vectorizer (HVec).


\subsection{Label Correction}

To address the issue of label noise for our task, we propose an unsupervised label correction method. We initially feed our word embeddings through a dimensionality reduction algorithm to convert the high-dimensional features outputted by the embedding models to lower-dimensional representation. Due to the nature of most clustering algorithms, high-dimensional data typically results in subpar performance and poorly separated clusters, a phenomenon known as the ``Curse of Dimensionality'' \cite{steinbach2004challenges}. Thus, representing the data in lower dimensions is a necessary procedure. We experiment with three separate dimensionality reduction algorithms: Principal Component Analysis (PCA), Deep Neural Autoencoders, and Uniform Manifold Approximation and Projection (UMAP) \cite{mcinnes2018umap}. PCA is a common reduction algorithm that extracts the most important information from a matrix of numerical data and represents it as a set of new orthogonal variables. Autoencoders use an unsupervised neural network to compress data into a low-dimensional space, and then reconstruct it while retaining the most possible information, enforcing efficient representation learning. The output of the encoder portion is used as the reduced embeddings. UMAP produces a graph from high-dimensional data and is optimized to generate a low-dimensional graph as similar to the input as possible. UMAP is specifically effective for high-dimensional data, as it has improved preservation of global structure and increased speed.

After reducing the dimensions of our word embeddings, we use clustering algorithms to separate them into two distinct clusters, which allow us to assign new labels to each post. We leverage clustering algorithms because of their unsupervised nature; this is critical because we have no prior knowledge regarding the noise distribution in the labels, requiring a clustering procedure which is independent of the web-scraped labels. We use the Gaussian Mixture Model (GMM) as our clustering algorithm. A GMM is a parametric probability density function used as a model of the probability distribution of continuous measurements in order to cluster given data using probabilities. As a baseline, we use K-Means clustering; K-means attempts to divide $n$ observations into $k$ clusters, such that each observation is assigned to the cluster with the closest mean, and the clusters minimize within-cluster distance while maximizing between-cluster distance. To avoid the dimensionality reduction requirement, we use subspace clustering via spectral clustering \cite{parsons2004subspace}, which specifically allows unsupervised clustering of high-dimensional data by identifying clusters in different sub-spaces within a dataset.

Each word embedding is now associated with two labels: the original labels based on the subreddit, which are the ground-truth labels, and the new labels resulting from unsupervised clustering. We subsequently leverage a confidence-based thresholding method to correct the ground-truth labels. If the clustering algorithm predicts a label with a probability above $\tau$, a tuned threshold, the ground-truth label is replaced with the predicted label; otherwise, we assume the ground-truth label. The tuned threshold ensures only predicted labels with high confidence are used to correct the ground-truth, preventing false corrections. Finally, the correct set of labels are paired with their respective post. We obtain class probabilities using the clustering algorithms. Note that our label correction method can be used in any NLP domain or even in other fields, such as the imaging field.

\subsection{Classification}

With a corrected label set, we train our deep neural networks to determine whether the posts display depressive or suicidal sentiment. Similar to the embedding process, any classifier can be used in place of the ones we tested. However, we aim to prove that deep neural classifiers are effective for our proposed task, as neural networks allow for accurate representation learning to differentiate the close semantics of our two classes.

For experimentation, we tested four deep learning algorithms: a dense neural network, a Convolutional Neural Network (CNN), a Bidirectional Long Short-Term Memory Neural Network (BiLSTM), a Gated Recurrent Unit Neural Network (GRU). For baselines, we evaluated three standard machine learning models: Logistic Regression (LogReg), Multinomial Naive Bayes (MNB), and a support-vector machine (SVM).

\subsection{Datasets}
\label{sub:datasets}

We develop a primary dataset based on our task of suicide or depression classification. This dataset is web-scraped from Reddit. We collect our data from subreddits using the Python Reddit API. We specifically scrape from two subreddits: r/SuicideWatch and r/Depression. The dataset contains 1,895 total posts. We utilize two fields from the scraped data: the original text of the post as our inputs, and the subreddit it belongs to as labels. Posts from r/SuicideWatch are labeled as suicidal, and posts from r/Depression are labeled as depressed. We make this dataset and the web-scraping script available in our code. 

Furthermore, we use the Reddit Suicide C-SSRS dataset \cite{gaur_manas_2019_2667859} to verify our label correction methodology. The C-SSRS dataset contains 500 Reddit posts from the subreddit r/depression. These posts are labeled by psychologists according to the Columbia Suicide Severity Rating Scale, which assigns progressive labels according to severity of depression. We use this dataset to validate our label correction method since the labels are clinically verified and from the same domain of Reddit. To further validate the label correction method, we use the IMDB large movie dataset, a commonly used NLP benchmark dataset \cite{maas-EtAl:2011:ACL-HLT2011}. The dataset is a binary classification task which contains 50,000 polar movie reviews. We use a random subset of samples for evaluation. 

For comparison of our method against other related tasks and methods, we build a dataset for binary classification of clinically healthy text vs suicidal text. We utilize the two subreddits r/CasualConversation and r/SuicideWatch. r/CasualConversation is a subreddit of general conversation, and has generally been used by other methods as data for a clinically healthy class \cite{schrading2015analysis}. 

\section{Experimental Results}
\label{sec:results}

\subsection{Implementation Details}

For all datasets, we set aside 20\% of the dataset as an external validation set. The deep learning models were implemented with Tensorflow, and the rest of the models were implemented with Sci-Kit Learn. We trained the deep learning models with the Adam optimizer and used a binary cross-entropy loss function. Based on tuning experiments, where we recorded accuracy at varying values, we set $\tau$ to 0.90 for all experiments, but similar values yielded similar performance. For classification accuracy, we use five metrics: Accuracy (Acc), Precision (Prec), Recall (Rec), F1-Score (F1), and Area Under Curve Score (AUC). Model-specific hyperparameters are included in the code.

\subsection{Label Correction Performance}

To evaluate our clustering performance, we present both the accuracy of the clustering algorithm at correcting noisy labels as well as classification performance after label correction. Classification on a clean test set is expected to decrease as training labels become noisier \cite{frenay2014survey}. Therefore, we contend that if after label correction, the classification accuracy of our algorithm increases, the correction method is effective. Importantly, because our proposed task uses a web-scraped dataset, the labels are not clinically verified. This unfortunately means evaluating the correction rate of noisy labels is impossible because we do not have the true labels. Therefore, we perform label correction evaluation on the benchmark IMDB dataset to demonstrate the value of our method in a general setting, as well as on the C-SSRS dataset to demonstrate effectiveness in our specific domain.

\subsubsection{Clustering Performance}

To evaluate the performance of clustering, we inject noise into the label set at different rates. We corrupt 10-40\% of the dataset at both uniform and imbalanced rates, as the noise rate of labels in real-world datasets are estimated to be 8\% to 38.5\%. These noise levels are also standard for other noisy label papers \cite{song2020learning}. We then evaluated the performance of the clustering algorithms at correcting the noisy labels.

\begin{figure}
    \centering
    \resizebox{0.8\linewidth}{!}{
    \begin{tabular}{cc}
    \includegraphics[width=1\linewidth]{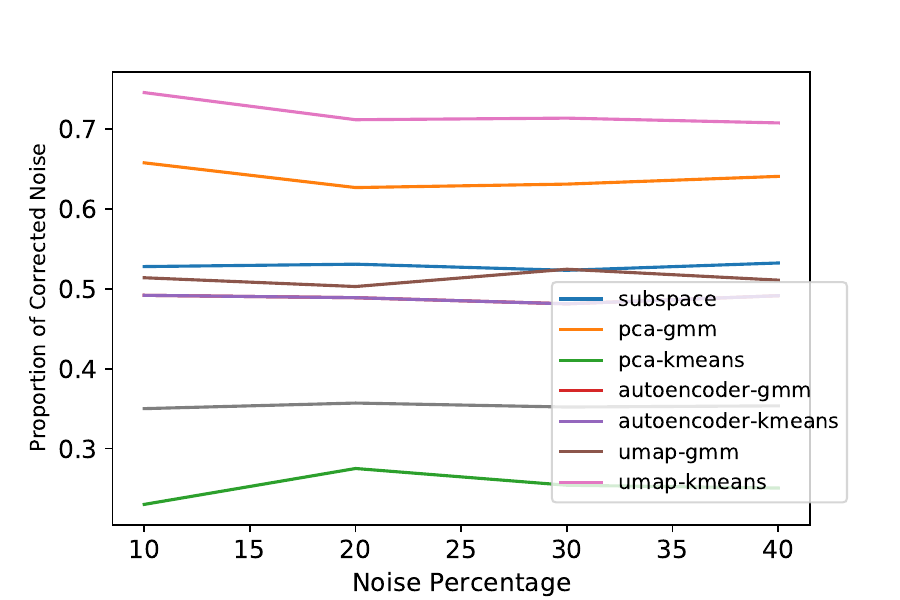} &
    \includegraphics[width=1\linewidth]{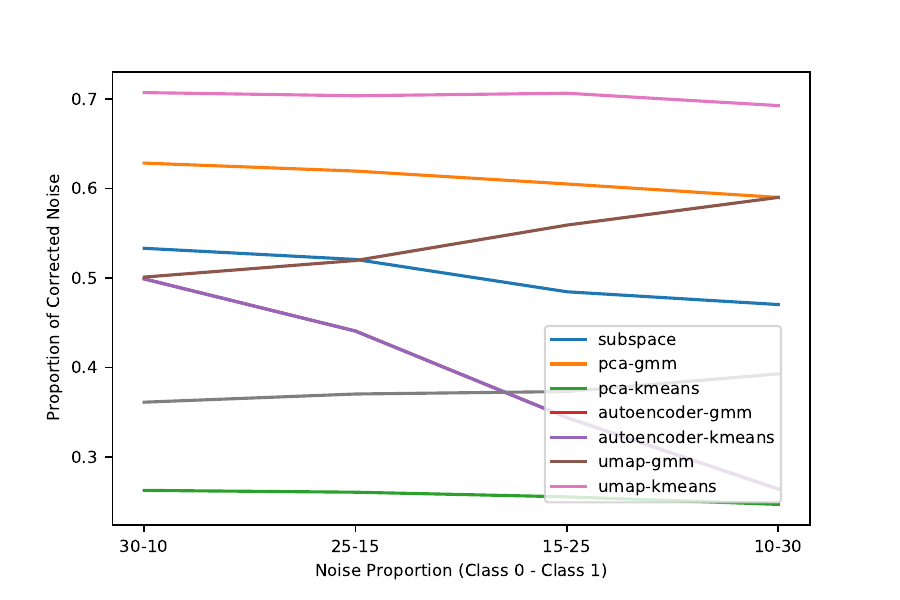}
    \\
    \includegraphics[width=\linewidth]{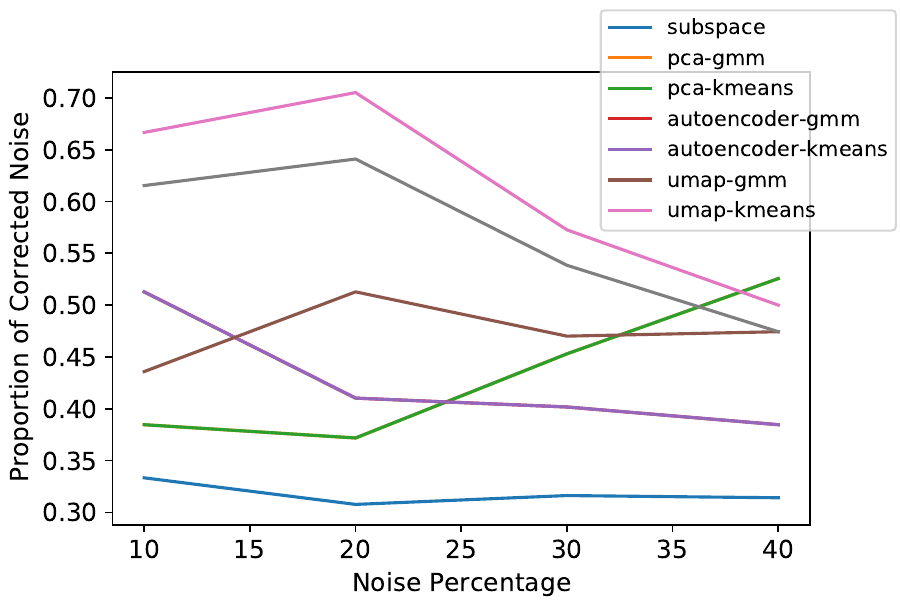} &
    \includegraphics[width=\linewidth]{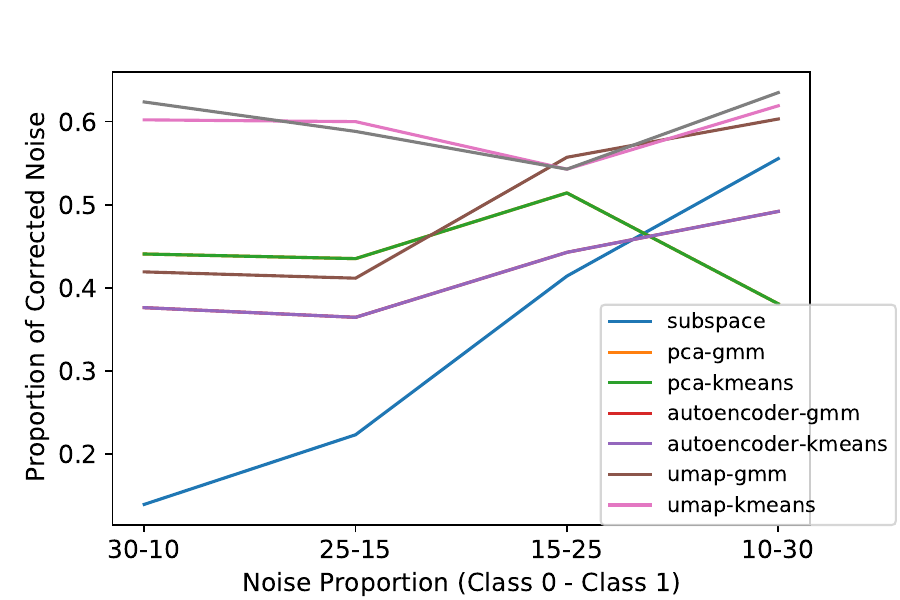}

    \end{tabular}
    }
    \caption{Correction rates of the label correction algorithms at different noise rates on the \textbf{IMDB} (top) and \textbf{C-SSRS} (bottom) dataset. Left: correction rates with uniform injections of noise. Right: correction rates with class-weighted injections of noise (ratios such as 30\%-10\% or 25\%-15\%).}
    \label{fig:correction}
\end{figure}



As seen in Figure \ref{fig:correction}, our noise correction method is able to consistently remove $>$ 50\% of injected noise while remaining below a 10\% false-correction rate on both datasets, and the performance does not degrade heavily at higher noise percentages, which is challenging to achieve \cite{frenay2014survey}. The SDCNL label correction is successful on both the IMDB dataset, which shows the generalizability of the method, as well as the C-SSRS dataset, proving its ability in our specific domain and task. The best combinations of reduction and clustering algorithms are umap-kmeans and umap-gmm, which we use as the proposed method. To our best knowledge, almost all noisy label correction methods do not evaluate correction rates but rather evaluate performance on classification accuracy after correction, as it allows for comparison to other noisy label methods that do not use label correction. However, because most recent noisy label methods are in the image domain, drawing comparisons to related work is unfeasible.

\subsubsection{Classification Performance after Label Correction}

Lastly, to demonstrate the effectiveness of the label correction method, we train a classifier on noisy C-SSRS data and validate on a separate C-SSRS test set which has no noise. We then use our label correction method to correct the same set of noisy labels, train the model with the correction labels, and validate on the same unmodified test set. 

\begin{table}
    \setlength{\tabcolsep}{4pt}
    \centering
    \resizebox{0.4\linewidth}{!}{
    \begin{tabular}{@{} c c ccc@{}}
        \toprule
        \multirow{2}{*}{Model} & \phantom{a} & \multicolumn{3}{c}{Accuracies per Task (\%)} \\
        \cmidrule{3-5}
        &&  Noisy && \textbf{Corrected} \\
        \midrule
        guse-dense &&  57.97 && 70.63 \\
        bert-dense  && 48.86 && 70.13 \\
        bert-bilstm  && 56.71 && 68.35 \\
        bert-cnn  && 55.70 && 70.13 \\
        \bottomrule
        \end{tabular}
    }
    \caption{Classifier accuracy comparison after injecting randomized noise (20\%) into C-SSRS labels (left) against using the label correction method (UMAP + GMM) to remove the artificial noise and subsequently training classifier (right).}
    \label{tab:noisy-corrected-scores}
\end{table}

\begin{figure}
    \centering
    \resizebox{0.6\linewidth}{!}{
    \includegraphics[width=\linewidth]{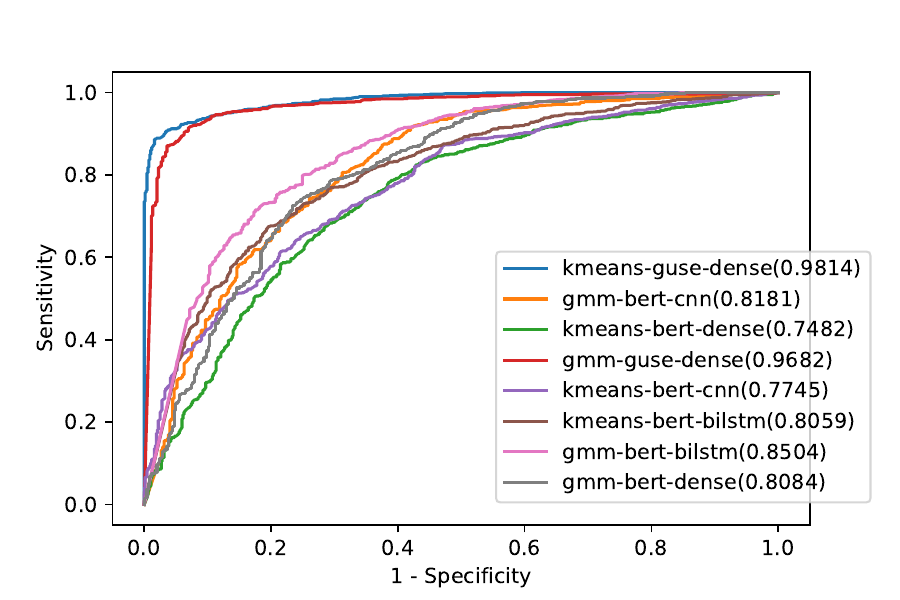}
    }
    \caption{ROC curves of model performance after using label correction. The 4 best combination of models with the two final label correction methods are shown (GMM vs K-Means). UMAP is used to reduce the dimensions of the embeddings.}
    \label{fig:roc-thresholding}
\end{figure}

We show that accuracy improves markedly after using our label correction method, as there is at least a 11\% increase (Table \ref{tab:noisy-corrected-scores}). Because our label correction process works on a dataset in the same domain, it is an effective method for cleaning noisy labels in NLP and for our task. Moreover, as seen in Figure \ref{fig:roc-thresholding}, using a probability threshold impacts performance, proving that using a threshold is an important factor in ensuring the corrected labels are accurate. Thus, we finalize on the thresholding method for our final model. 

\subsection{Classification Performance}

\subsubsection{Deep Neural Network Performance}

\begin{table}
    \setlength{\tabcolsep}{4pt}
    \centering
    \resizebox{0.6\linewidth}{!}{
    \begin{tabular}{cc cc cc cc cc}
        \toprule
        \multirow{2}{*}{Metrics (\%)} & \phantom{a} & \phantom{a} & \multicolumn{4}{c}{Model Combinations} \\
        \cmidrule{3-9}
        &&  guse-dense && bert-dense && bert-bilstm && bert-cnn\\
        \midrule
        Acc &&  \textbf{72.24} && 70.50 &&  71.50 && 72.14 \\
        Rec  && \textbf{76.37} && 71.92 &&  67.77 && 73.99 \\
        Prec  && 71.38 && 70.77 &&  \textbf{74.28} && 72.18 \\
        F1  && \textbf{73.61} && 71.25 &&  70.70 && 72.92 \\
        AUC  && \textbf{77.76} && 75.43 &&  77.11 && 76.35 \\
        \bottomrule
        \end{tabular}
    }
    \caption{Performance of the four best combinations of embedding models and classifiers.}
    \label{tab:4-best-models}
\end{table}

After performing all experiments, we determined the four strongest combinations to perform the remainder of the tests. These combinations are trained on the primary suicide vs depression dataset with uncorrected labels. The complete results are in Appendix A in the supplemental (in Github repository). The performance of the four strongest models are shown in Table \ref{tab:4-best-models}. The combinations are: BERT embeddings with a CNN (bert-cnn), BERT with a fully-dense neural network (bert-dense), BERT with a Bi-LSTM neural network (bert-bilstm), and GUSE with a fully-dense neural network (guse-dense). This proves the importance of our contribution as all DNNs outperform the baselines. For all future experiments, we use the four models above. 

\subsubsection{Comparison to Other Tasks}

While there is extensive research on NLP text-based approaches to suicide detection, there is none for our specific task of low-risk depression versus suicidal ideation. We performed an additional test of our proposed model by testing the standard task of classifying suicide versus clinically healthy.


\begin{figure}
    \centering
    \begin{minipage}{.4\textwidth}
      \centering
    \includegraphics[width=\linewidth]{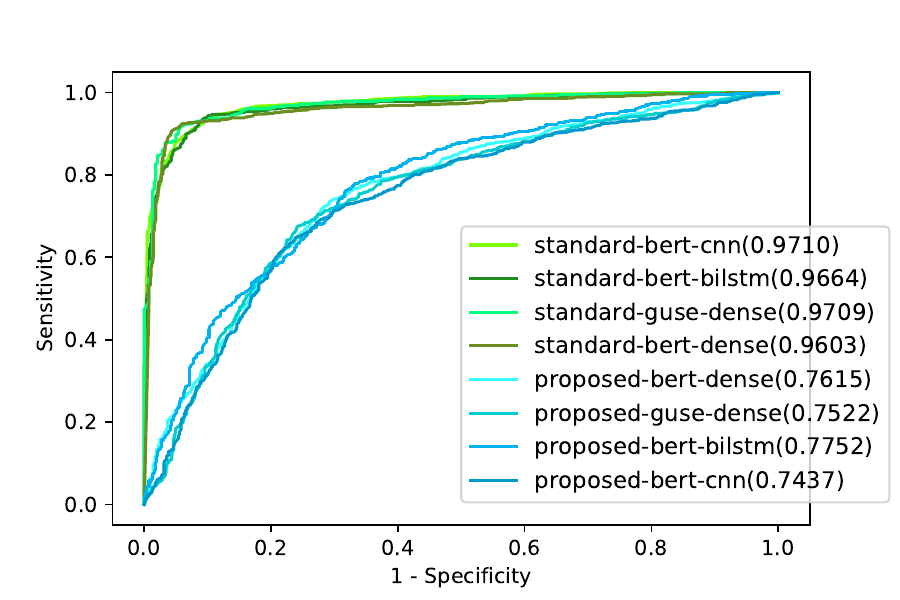}
    \caption{ROC curves of model performance from four best models on our task (proposed) against the conventional suicide vs healthy task (standard).} 
    \label{fig:roc-easy-final}
    \end{minipage}%
    \hspace{0.2cm}
    \begin{minipage}{.4\textwidth}
      \centering
    \includegraphics[width=\linewidth]{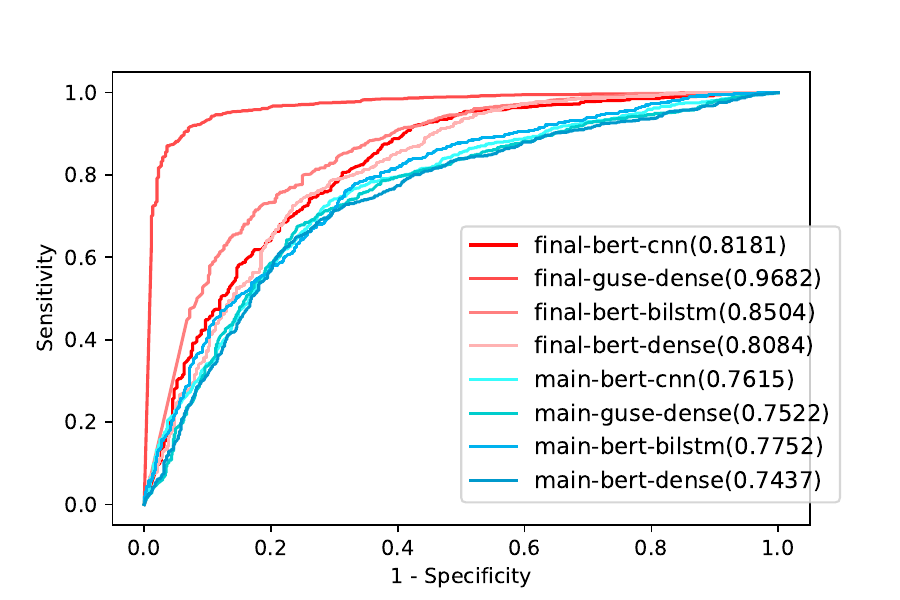}
    \caption{ROC curves of performance of top 4 models with label correction (red) against the same models without using label correction (blue).}
    \label{fig:roc-final}
    \end{minipage}
\end{figure}

Figure \ref{fig:roc-easy-final} displays that on identical models, the commonly-researched task achieved much stronger baseline performance compared to our task; therefore, our baseline evaluations should correspondingly be lower. This task difficulty also demonstrates the value of automated methods such as ours in clinical settings.

\subsubsection{Final Evaluation}

We evaluated the classification performance of SDCNL on the Reddit dataset. We generated ground-truth labels for this dataset using the proposed label correction method. To provide a complete test of our model, it would be preferable to use labels provided by a mental health professional; however, no such dataset exists for our task. We contend that, since we demonstrated that the proposed label correction method is effective on the C-SSRS and IMDB datasets, the use of the label correction method on the Reddit dataset is justified.

\begin{table}
    \setlength{\tabcolsep}{4pt}
    \centering
    \resizebox{\linewidth}{!}{
    \begin{tabular}{@{} c c cccc c cccc@{}}
    \toprule
    \multirow{2}{*}{Metrics (\%)} & \phantom{a} & \multicolumn{4}{c}{UMAP-KMeans} & \phantom{a} & \multicolumn{4}{c}{UMAP-GMM}\\
    \cmidrule{3-6}\cmidrule{8-11}
    && guse-dense & bert-dense & bert-bilstm & bert-cnn && \textbf{\textit{guse-dense}} & bert-dense & bert-bilstm & bert-cnn \\
    \midrule
    Acc &&  \textbf{92.61} & 73.56 &  75.15 & 74.72 && \textbf{93.08} & 83.74 & 84.16 & 84.59 \\
    Prec  && \textbf{93.61} & 83.18 & 84.02 & 87.38 && 94.76 & \textbf{95.51} & 93.10 & 95.38 \\
    Rec  && \textbf{94.85} & 77.66 & 79.00 & 76.65 && \textbf{96.16} & 85.08 & 87.09 & 86.05 \\
    F1  && \textbf{94.22} & 80.25 & 81.19 & 81.64 && \textbf{95.44} & 89.99 & 89.99 & 90.45 \\
    AUC  && \textbf{98.18} & 76.83 & 80.93 & 78.69 && \textbf{96.88} & 81.97 & 85.08 & 82.91 \\
    \bottomrule
    \end{tabular}
    }
    \caption{Final classification performance after using the two label correction methods, with and without a thresholding scheme. Best performances for each noise removal method are bolded. Best overall model is bolded and italicized.}
    \label{tab:final-scores}
\end{table}


Table \ref{tab:final-scores} displays the final performance of the models with both threshold label correction (GMM) and without (K-Means). We validate the importance of the thresholding component, as all metrics are substantially improved over the K-Means baseline. Moreover, as displayed in Figure \ref{fig:roc-final}, our label correction method improves the ROC curve and yields a much higher AUC value. By correcting the labels in the test set with high expected noise, we achieve substantially higher performance. Our final proposed combination uses GUSE as the embedding model, a fully-dense network for the classifier, and corrects labels using UMAP for dimensionality reduction and a GMM for the clustering algorithm.
GUSE embeddings likely yield the best results due to outputting less embeddings than the other transformers, preserving information. 

\section{Conclusions and Ethical Discussion}

In this paper, we present SDCNL, a novel method for deep neural network classification of depressive sentiment vs suicidal ideation with unsupervised noisy label correction. The use of deep neural networks allows for effective classification of closely related classes on a proven difficult task. Our novel method of label correction using unsupervised clustering effectively removes high-volumes of noise from both benchmark and domain-specific datasets, allowing for the use of large-scale, web-scraped datasets. Our extensive experimentation and ablative results highlight the effectiveness of our proposed model and its potential for real diagnostic application. 

The applied setting of our system is to provide professionals with a supplementary tool for individual patient diagnosis, as opposed to solely being a screening method on social media platforms. SDCNL could be used by professional therapists as a ``second opinion'', friends and family as a preliminary screening for loved ones, or even on social media platforms to identify at-risk users.

This paper is not a clinical study, and the results are suitable for research purposes only. Were our algorithm to be used as a diagnostic tool, the main ethical concern would be false negative and false positive predictions. Specifically when dealing with suicide, which is a life or death situation, AI systems alone are not sufficient to provide proper screening. Future researchers who work on our topic or with our paper must be aware of these ethical concerns and not make major steps without proper clinical support. This paper is solely meant to demonstrate the potential efficacy of a suicide prevention mechanism.

Throughout this study, we collected our data while protecting user privacy and maintaining ethical practices. We de-identified our dataset, which we made publicly available, by removing personal information such as usernames. Moreover, Reddit is a public and anonymous forum, meaning our data source was anonymized and in the public domain to begin with.



\bibliographystyle{splncs04}
\bibliography{references.bib}

\appendix
\section{Baseline Model Evaluations}
\label{appendix:A}

Appendix \ref{appendix:A} displays the full experimentation of baseline results for 6 transformers and 7 classification algorithms (Table \ref{table:all-scores}). The results are evaluated with 5 metrics: Accuracy (Acc), Precision (Prec), Recall (Rec), F1 score (F1), and Area Under Curve (AUC). With these results, we selected the four strongest performing combinations of models to perform the remainder of the experimentation. 

\begin{table}[h]
\centering
\setlength{\tabcolsep}{4pt}
\medskip
\label{table:all-scores}
\resizebox{0.82\linewidth}{!}{
\begin{tabular}{@{} lc c ccccccc@{}}
            \toprule
          \multirow{3}{*}{Embedding Model}
          &
           \phantom{a}
           &
           \phantom{a}
           &
           \phantom{a}
           &
          \multicolumn{5}{c}{Classifiers}
           \\
           \cmidrule{4-10}
          && Metrics & CNN & Dense & BiLSTM & GRU & MNB & SVM & LogReg
          \\
          \midrule
            \multirow{5}{*}{BERT} 
           &&
           Acc & 72.14 & 70.50 & 71.50 & 71.50 & 57.78 & 68.07 & 68.60
           \\
           &&
           Rec & 73.99 & 71.92 & 67.78 & 68.91 & 53.37 & 73.58 & 70.98
           \\
           &&
           Prec & 72.18 & 70.77 & 74.28 & 73.86 & 59.54 & 66.98 & 68.50
           \\
           &&
           F1 & 72.92 & 71.25 & 70.70 & 71.05 & 56.28 & 70.12 & 69.72
           \\
           &&
           AUC & 76.35 & 75.43 & 77.11 & 76.66 & 54.21 & 55.43 & 54.72
           \\
          \midrule
          \multirow{5}{*}{SentenceBERT} 
           &&
           Acc & 68.65 & 68.87 & 69.55 & 70.77 & 59.37 & 68.34 & 63.85
           \\
           &&
           Rec & 73.37 & 74.61 & 67.98 & 67.36 & 46.63 & 73.06 & 69.95
           \\
           &&
           Prec & 67.88 & 67.94 & 71.22 & 73.35 & 63.83 & 67.46 & 63.08
           \\
           &&
           F1 & 70.40 & 70.82 & 69.41 & 70.01 & 53.89 & 70.15 & 66.34
           \\
           &&
           AUC & 73.52 & 73.70 & 74.00 & 74.99 & 56.13 & 53.12 & 51.33
           \\
           \midrule
           \multirow{5}{*}{GUSE} 
           &&
           Acc & 72.66 & 72.24 & 72.82 & 73.19 & 69.39 & 71.50 & 71.50
           \\
           &&
           Rec & 78.96 & 76.37 & 78.03 & 77.10 & 67.36 & 75.65 & 74.61
           \\
           &&
           Prec & 70.79 & 71.38 & 71.36 & 72.31 & 71.04 & 70.53 & 70.94
           \\
           &&
           F1 & 74.62 & 73.61 & 74.49 & 74.52 & 69.15 & 73.00 & 72.73
           \\
           &&
           AUC & 77.82 & 77.76 & 77.41 & 76.33 & 47.68 & 49.47 & 50.75
           \\
           \midrule
          \multirow{5}{*}{TFIDF} 
           &&
           Acc & 67.12 & 69.28 & 67.97 & 67.92 & 69.39 & 70.45 & 69.13
           \\
           &&
           Rec & 81.35 & 67.78 & 72.33 & 74.61 & 76.68 & 74.09 & 73.58
           \\
           &&
           Prec & 65.98 & 70.73 & 67.41 & 66.61 & 67.58 & 69.76 & 68.27
           \\
           &&
           F1 & 71.89 & 69.19 & 69.64 & 70.21 & 71.85 & 71.86 & 70.83
           \\
           &&
           AUC & 70.04 & 75.23 & 71.70 & 72.23 & 51.65 & 51.38 & 51.35
           \\
           \midrule
           \multirow{5}{*}{CountVec} 
           &&
           Acc & 69.18 & 68.92 & 68.13 & 67.86 & 66.75 & 65.43 & 63.32
           \\
           &&
           Rec & 82.07 & 73.47 & 75.23 & 74.82 & 72.54 & 69.43 & 66.84
           \\
           &&
           Prec & 65.91 & 68.32 & 66.62 & 66.41 & 65.73 & 65.05 & 63.24
           \\
           &&
           F1 & 73.06 & 70.61 & 70.59 & 70.31 & 68.97 & 67.17 & 64.99
           \\
           &&
           AUC & 74.44 & 73.34 & 72.66 & 72.30 & 51.47 & 51.08 & 47.35
           \\
           \midrule
           \multirow{5}{*}{HashVec} 
           &&
           Acc & 67.86 & 67.44 & 65.49 & 65.17 & 65.96 & 66.23 & 66.75
           \\
           &&
           Rec & 71.50 & 69.02 & 66.74 & 68.19 & 69.43 & 69.95 & 72.54
           \\
           &&
           Prec & 67.58 & 67.79 & 66.34 & 65.30 & 65.69 & 65.85 & 65.73
           \\
           &&
           F1 & 69.27 & 68.31 & 66.17 & 66.47 & 67.51 & 67.84 & 68.96
           \\
           &&
           AUC & 71.47 & 71.63 & 68.98 & 69.58 & 52.94 & 54.06 & 52.85
           \\
           \bottomrule
          \end{tabular}}
          \caption{Performance of all 42 combinations. 7 classifiers and 6 embedding models are used, with 4 deep classifiers and 3 deep embedding models.}
\end{table}

\section{Comparison to Conventional Task}
\label{appendix:B}

Appendix \ref{appendix:B} presents an additional table showing the performance of our 4 best models on both the conventional task of suicidal vs clinically healthy classification against our task of suicidal vs depression classification. The results demonstrate the difficulty of our task and justify our clinical motivations.

\begin{table}[h]
    \setlength{\tabcolsep}{4pt}
    \centering
    \resizebox{\linewidth}{!}{
    \begin{tabular}{@{} c c cccc c cccc@{}}
    \toprule
    \multirow{2}{*}{Metrics (\%)} & \phantom{a} & \multicolumn{4}{c}{Proposed} & \phantom{a} & \multicolumn{4}{c}{\textbf{Standard}}\\
    \cmidrule{3-6}\cmidrule{8-11}
    && guse-dense & bert-dense & bert-bilstm & bert-cnn && guse-dense & bert-dense & bert-bilstm & bert-cnn \\
    \midrule
    Acc &&  72.24 & 70.50 &  71.50 & 72.14 && 92.28 & 57.46 & 92.78 & 93.11 \\
    Prec  && 76.37 & 71.92 & 67.77 & 73.99 && 93.56 & 80.70 & 92.16 & 92.26 \\
    Rec  && 71.38 & 70.77 & 74.28 & 72.18 && 92.46 & 58.28 & 94.54 & 95.07 \\
    F1  && 73.61 & 71.25 & 70.70 & 72.92 && 93.00 & 67.39 & 93.35 & 93.63 \\
    AUC  && 77.76 & 75.43 & 77.11 & 76.35 && 96.65 & 54.91 & 97.24 & 96.49 \\
    \bottomrule
    \end{tabular}
    }
    \caption{Comparison of classification metrics between conventionally researched task of suicide vs clinically healthy against our proposed task of suicide vs depression. The better performing category is bolded. The standard task performs far better on the same models, highlighting how our proposed task is more difficult to categorize.}
    \label{tab:easy-main-scores}
\end{table}

\section{Vectorizers and Clustering}
\label{appendix:C}

\begin{figure}[h]
    \centering
    \resizebox{0.9\linewidth}{!}{
    \includegraphics[width=0.49\linewidth]{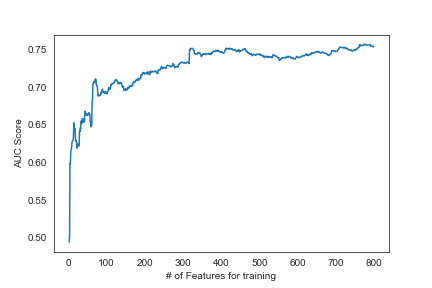}
    \hfill
    \includegraphics[width=0.49\linewidth]{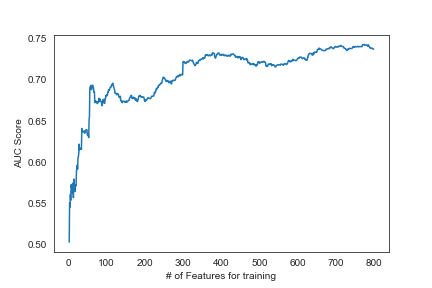}
    \hfill
    \includegraphics[width=0.49\linewidth]{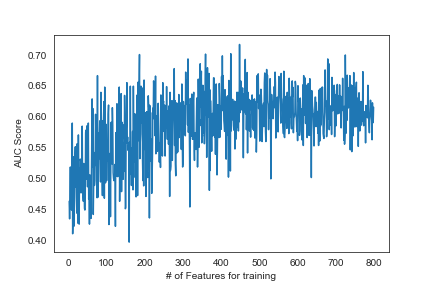}
    }
    \caption{Area Under Curve (AUC) values at different number of extracted word embeddings/features from the three vectorizers (TFIDF, CVec, HVec) inputted into Bayesian classifier. AUC plateaus at 
    around 400 features, so we used 768 features to be consistent with other word embedding models.}
    \label{fig:features-vs-auc}
\end{figure}

\begin{figure}[h]
    \centering
    \resizebox{0.5\linewidth}{!}{
    \includegraphics[width=0.49\linewidth]{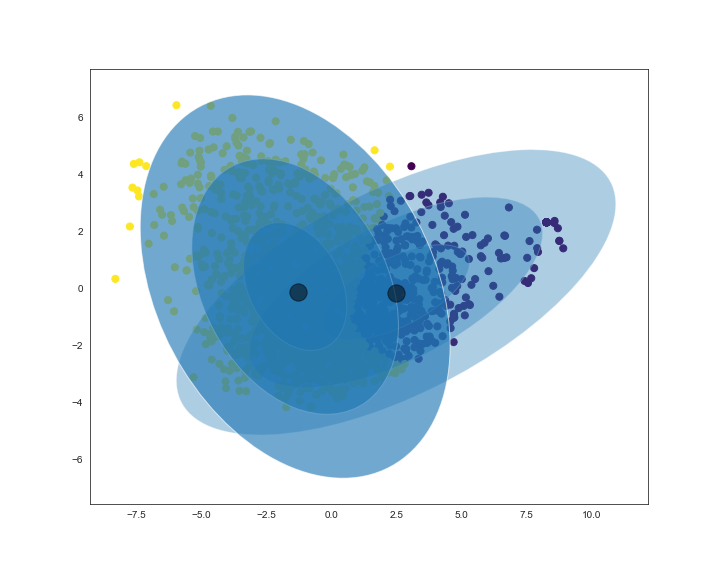}
    }
    \caption{GMM clustering using BERT embeddings and PCA reduction to 2 dimensions show the difficulty of the clustering task, as there is little variety in the clusters and they heavily overlap. We use co-variance type "full".}
    \label{fig:gmm-plot}
\end{figure}

Appendix \ref{appendix:C} shows how the number of extracted features from the vectorizers was chosen (Figure \ref{fig:features-vs-auc}). The AUC scores from the MNB classifier after using the vectorizers with different number of embeddings are shown. After around 400 features for each model, the performance converges, so we finalized on extracting 768 features for consistency. Figure \ref{fig:gmm-plot} visualizes the clustering of a GMM using BERT embeddings with PCA reduction. 

\end{document}